\documentclass[conference]{IEEEtran}
\IEEEoverridecommandlockouts

\usepackage{cite}
\usepackage{amsmath,amssymb,amsfonts}
\usepackage{algorithmic}
\usepackage{graphicx}
\usepackage{textcomp}
\usepackage{xcolor}
\def\BibTeX{{\rm B\kern-.05em{\sc i\kern-.025em b}\kern-.08em
    T\kern-.1667em\lower.7ex\hbox{E}\kern-.125emX}}

\usepackage{orcidlink}
\usepackage[linesnumbered,commentsnumbered,ruled,lined]{algorithm2e}
\usepackage{tabularx}
\usepackage{adjustbox}
\usepackage{booktabs}
\usepackage{hyperref}
\usepackage{multirow}
\usepackage{makecell}

\begin{document}

\title{Enhancing Multi-Attribute Fairness in Healthcare Predictive Modeling\\
}

\author{
\IEEEauthorblockN{Xiaoyang Wang}
\IEEEauthorblockA{\textit{College of Computing and Informatics} \\
\textit{Drexel University}\\
Philadelphia, USA \\
xw388@drexel.edu}
\and
\IEEEauthorblockN{Christopher C. Yang}
\IEEEauthorblockA{\textit{College of Computing and Informatics} \\
\textit{Drexel University}\\
Philadelphia, USA \\
chris.yang@drexel.edu}
}

\maketitle

\begin{abstract}
Artificial intelligence (AI) systems in healthcare have demonstrated remarkable potential to improve patient outcomes.  However, if not designed with fairness in mind, they also carry the risks of perpetuating or exacerbating existing health disparities. Although numerous fairness-enhancing techniques have been proposed, most focus on a single sensitive attribute and neglect the broader impact that optimizing fairness for one attribute may have on the fairness of other sensitive attributes. In this work, we introduce a novel approach to multi-attribute fairness optimization in healthcare AI, tackling fairness concerns across multiple demographic attributes concurrently.  Our method follows a two-phase approach: initially optimizing for predictive performance, followed by fine-tuning to achieve fairness across multiple sensitive attributes. We develop our proposed method using two strategies, sequential and simultaneous. Our results show a significant reduction in Equalized Odds Disparity (EOD) for multiple attributes, while maintaining high predictive accuracy. Notably, we demonstrate that single-attribute fairness methods can inadvertently increase disparities in non-targeted attributes whereas simultaneous multi-attribute optimization achieves more balanced fairness improvements across all attributes. These findings highlight the importance of comprehensive fairness strategies in healthcare AI and offer promising directions for future research in this critical area.
\end{abstract}


\section{Introduction}
The rapid growth in data availability and computational capabilities has significantly enhanced the efficacy of machine learning techniques. Consequently, these algorithms have become integral to automated decision-making processes across diverse real-world domains. Specifically, Artificial Intelligence (AI) has emerged as a powerful tool in healthcare, promising to revolutionize diagnosis, treatment planning, and patient care. However, the increasing adoption of AI in healthcare has raised significant concerns about fairness and equity, particularly when these systems make decisions that affect diverse patient populations~\cite{mehrabi_survey_2021}. AI models trained on historical data may inadvertently perpetuate or even exacerbate existing biases, leading to disparities in healthcare outcomes across different demographic groups. This challenge is particularly acute in healthcare, where factors such as race, gender, age, and socioeconomic status can significantly influence both health status and access to care.

Many studies have been conducted to assess bias in predictive modeling and enhance fairness through a variety of methodological interventions. The strategies employed to mitigate bias and promote fairness in machine learning models can be classified into three categories, pre-processing, in-processing, and post-processing, corresponding to specific stages of the model development process~\cite{friedler2019comparative}. While numerous fairness-enhancing techniques have been proposed in recent years, most focus on addressing bias with respect to a single sensitive attribute, such as race or gender. However, real-world healthcare scenarios often involve multiple, intersecting demographic factors that can contribute to unfair outcomes. The complexity of these intersectional fairness issues necessitates more sophisticated approaches that can simultaneously address multiple dimensions of demographic diversity~\cite{padhAddressingFairnessClassification2021}. There is a pressing need for methods to enhance fairness across multiple sensitive attributes without significantly compromising the predictive performance of AI models in critical healthcare applications.

To address this challenge, we propose a method based on transfer learning to enhance fairness for multiple demographic groups in healthcare AI systems. Our approach consists of two primary phases: first, we optimize the model for maximum predictive performance, and then we transfer this performance-optimized model to a fairness optimization phase. During the fairness optimization, we employ a carefully designed loss function coupled with a penalty term to improve fairness across multiple demographic attributes while maintaining the model's predictive capabilities. We explore this method through two strategies: a sequential approach that optimizes fairness for one attribute at a time, and a simultaneous approach that addresses multiple attributes simultaneously.

The key contributions of this work are threefold. First, we introduce a transfer learning-based framework that effectively balances the dual objectives of predictive performance and multi-attribute fairness in healthcare AI. Second, we provide empirical evidence of our method's effectiveness using two real-world healthcare datasets, demonstrating significant fairness improvements across multiple attributes. Finally, we offer insights into the trade-offs between sequential and simultaneous fairness optimization strategies, revealing that sequential strategy tends to favor the first-optimized attribute, while simultaneous strategy achieves more balanced fairness improvements across attributes. These findings have important implications for the design and deployment of fair AI systems in healthcare, particularly in contexts where multiple dimensions of demographic fairness must be considered.

\section{Preliminary}
In this section, we delineate the key notations employed throughout this study with Table~\ref{tab:symbols}.

\begin{table}[htbp]
\centering
\caption{Table of Symbols}
\label{tab:symbols}
\begin{tabular}{cl}
\toprule
\textbf{Symbol} & \textbf{Definition} \\
\midrule
$\mathcal{D}$ & The set of data points \\
$X \in \mathbb{R}^n$ & Feature vector of a data point \\
$Y \in \{0,1\}$ & Actual binary outcome \\
$\hat{Y} \in \{0,1\}$ & Model's predicted binary outcome \\
$Z$ & Sensitive attribute of the data point \\
$\mathcal{M}$ & Predictive model \\
$\mathbf{f}$ & Function implemented by $\mathcal{M}$ \\
$\boldsymbol{\theta}$ & Parameters of predictive model \\
\bottomrule
\end{tabular}
\end{table}

In this study, we assumed the sensitive attribute \( Z \) as a binary variable such as sex (where 0 signifies male and 1 denotes female) or racial identification (where 0 indicates Non-Caucasian and 1 represents Caucasian). We define subsets of $\mathcal{D}$ based on these attributes. For instance, the set of true positive cases for $Z=1$ is denoted as:
\[
\mathcal{D}_{Z=1, Y=1, \hat{Y}=1} = \{ (X, Y, \hat{Y}) \in \mathcal{D} \mid Z = 1, Y = 1, \hat{Y} = 1 \}
\]

\section{Related Work}

\subsection{Group Fairness in Machine Learning}

Group fairness in machine learning aims to ensure that protected groups, defined by sensitive attributes such as race, sex, or age, receive equitable treatment or outcomes from algorithmic decisions. This concept has gained significant attention, particularly in high-stakes domains like healthcare, where biased decisions can have severe consequences \cite{rajkomar2018ensuring}.

\subsubsection{Demographic Parity}
One of the earliest and most intuitive notions of group fairness is \textit{demographic parity} \cite{dwork2012fairness}. This criterion requires that the probability of a positive prediction is the same across all groups defined by the sensitive attribute $Z$. Formally, for a binary classifier $f$, demographic parity is satisfied if:

\begin{equation}
P(\hat{Y} = 1 \mid Z = a) = P(\hat{Y} = 1 \mid Z = b), \quad \forall a, b \in Z
\end{equation}

While intuitive, demographic parity can conflict with accuracy, especially when base rates differ between groups~\cite{hardt2016equality}. In healthcare, enforcing demographic parity without considering underlying differences in disease prevalence may lead to suboptimal outcomes.

\subsubsection{Equalized Odds and Equal Opportunity}
To address the limitations of demographic parity, Hardt et al. \cite{hardt2016equality} proposed the notions of \textit{Equalized Odds} and \textit{Equal Opportunity}. Equalized Odds requires equal true positive rates and false positive rates across all protected groups:

\begin{align}
P(\hat{Y} = 1 | Z = a, Y = y) &= P(\hat{Y} = 1 | Z = b, Y = y), \nonumber \\
&\forall a, b \in Z, y \in \{0, 1\}
\end{align}

Equal opportunity is a relaxation of equalized odds, requiring only equal true positive rates. 
\begin{align}
P(\hat{Y} = 1 | Z = a, Y = 1) &= P(\hat{Y} = 1 | Z = b, Y = 1), \nonumber \\
&\forall a, b \in Z, y \in \{0, 1\}
\end{align}
These metrics have been widely adopted in various domains, including healthcare predictive modeling \cite{pfohl2019creating}.

\subsubsection{Calibration}
Another important fairness criterion, especially relevant in risk prediction tasks common in healthcare, is calibration \cite{kleinberg2016inherent}. A model is well-calibrated with respect to protected groups if, for any predicted probability $p$, the fraction of positive outcomes in each group receiving this prediction is approximately $p$. Formally:

\begin{equation}
P(Y = 1 | f(X) = p, Z = z) = p, \quad \forall p \in [0,1], z \in \{a, b\}
\end{equation}

Calibration is crucial in healthcare applications where risk scores directly inform clinical decisions \cite{zink2020fair}.

\subsection{Bias Mitigation}

Approaches to mitigate bias in machine learning models can be categorized into three main strategies:

\begin{itemize}
    \item \textbf{Pre-Processing}: Pre-processing techniques modify the training data to remove biases before model training. Methods include reweighing \cite{kamiran2012data, peng2022fairmask}, resampling~\cite{lucas_resampling_2023}, and debiasing word embeddings \cite{bolukbasi2016man} for natural language processing tasks. In healthcare, Cerrato et al.~\cite{cerrato2020constraining} proposed a method to constrain the latent space of auto-encoders, removing sensitive information from patient data representations to prevent biased predictions.

    \item \textbf{In-Processing}: In-processing approaches involve modifying the learning algorithm to account for fairness during model training. Regularized optimization integrates fairness constraints directly into the model's objective function, augmenting the traditional loss function with a term that penalizes disparities across protected groups~\cite{kamishima_fairnessaware_2012, agarwal_reductions_2018, shen_optimising_2022, chang2024explainable}. Adversarial debiasing~\cite{zhang_mitigating_2018} and fair representation learning~\cite{zemel2013learning} are also prominent examples. In the healthcare domain, Pfohl et al.~\cite{pfohl2019creating} developed an adversarial approach to learn fair representations of clinical data, aiming to reduce bias while preserving predictive performance.

    \item \textbf{Post-Processing}: Post-processing techniques adjust the outputs of a trained model to ensure fairness without altering the model itself. Hardt et al.~\cite{hardt2016equality} proposed a method to achieve Equal Opportunity by modifying the decision thresholds for different groups. Fish et al.~\cite{fish2016confidence} introduced a classification paradigm based on confidence thresholds, assigning positive classifications only when predictive confidence exceeds a certain value. In healthcare, Zink and Rose~\cite{zink2020fair} developed a post-processing method to ensure fair risk predictions across different demographic groups in clinical decision support systems.
\end{itemize}

\subsection{Fairness in Healthcare AI}
In the context of healthcare, group fairness takes on added complexity due to the inherent differences in health conditions and outcomes across demographic groups. Rajkomar et al. \cite{rajkomar2018ensuring} discuss the challenges of implementing fairness in clinical predictive models, highlighting the need for careful consideration of the clinical context when defining and measuring fairness.

Chen et al. \cite{chen2019can} explored the tension between different notions of fairness in clinical risk prediction models, demonstrating that optimizing for one fairness metric often comes at the cost of others. This underscores the need for domain-specific approaches to fairness in healthcare AI.

While these studies have significantly advanced the understanding of fairness in machine learning and healthcare AI, they predominantly address bias concerning a single sensitive attribute. However, patients often belong to multiple protected groups simultaneously, and biases can intersect in complex ways~\cite{crenshaw2013demarginalizing}. Our work extends beyond this limitation by addressing fairness across multiple demographic groups. Many existing fairness optimization methods, such as removing certain sensitive information~\cite{chen2019fairness,cerrato2020constraining} or setting different thresholds for different groups~\cite{zink2020fair}, may be unsuitable for healthcare scenarios. These approaches can compromise diagnostic accuracy, introduce inconsistencies in clinical decision-making, and reduce the overall effectiveness of models. In healthcare AI, maintaining complete patient data integrity and ensuring consistent decision processes across all demographic groups is crucial for both ethical and clinical reasons. In contrast, our method, as an in-processing approach, adds fairness interventions during model training, aiming to maintain optimal predictive performance while enhancing fairness across multiple attributes. By leveraging a two-phase approach—first optimizing for performance and then fine-tuning for fairness—we strive to achieve a more balanced and practical solution for real-world healthcare applications.

\section{Methods and Materials}

\begin{figure*}[t]
    \centering
    \includegraphics[width=1\linewidth]{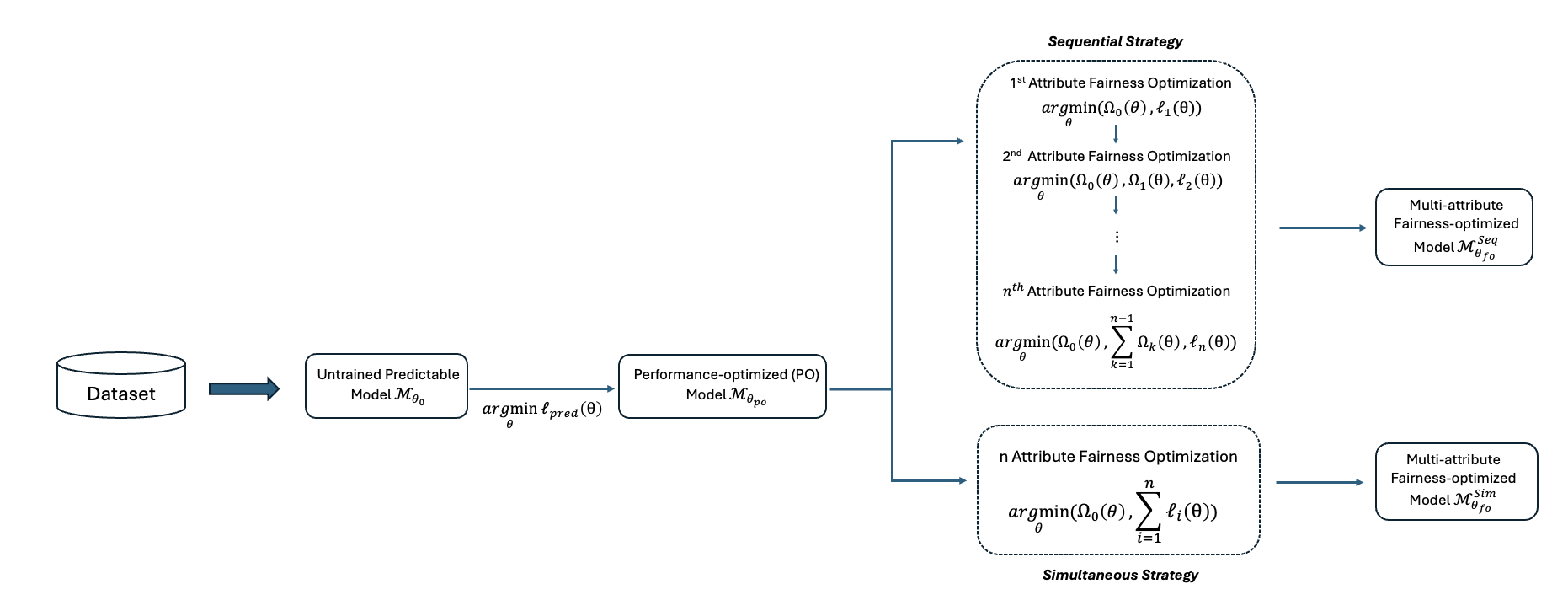}
    \caption{The multi-attribute fairness optimization pipeline, illustrating the performance optimization phase followed by the fairness optimization phase using sequential and simultaneous strategies.}
    \label{pipeline}
\end{figure*}
Fig~\ref{pipeline} presents our proposed methodology for developing fair and effective healthcare AI models. The Model Development phase includes performance optimization and fairness optimization. We explore two strategies for fairness optimization: \textit{Sequential}, which addresses fairness attributes one by one, and \textit{Simultaneous}, which optimizes all fairness attributes simultaneously.

\subsection{Performance Optimization}
The first phase of our proposed method focuses on optimizing the model for the best predictive performance. This serves as the foundation for subsequent fairness enhancements. In this study, we choose the logistic regression model because of its high interpretability and convergence stability, which are valuable in healthcare applications.

Let $\mathcal{D} = \{(\mathbf{x}_i, z_i, y_i)\}_{i=1}^n$ be the dataset, where $x_i \in \mathbb{R}^d$ represents the feature vector, $z_i$ the sensitive attributes (e.g., race and sex), and $y_i \in \{0,1\}$ the binary target variable for the $i$-th instance. We denote our predictive model as $\mathcal{M}_{\theta}(x)$, parameterized by $\theta$.

The optimization problem of this phase can be formulated to find the optimal parameters $\theta_{po}$ that minimize the prediction loss:
\begin{equation}
\theta_{po} = \arg\min_{\theta} \ell_\text{pred}(\theta)
\end{equation}
where $\ell_\text{pred}$ is the binary cross-entropy as our performance loss function, which is defined as:
\begin{equation}
\ell_\text{pred}(\theta) = -\frac{1}{n} \sum_{i=1}^n [y_i \log(\mathcal{M}_\theta(x_i)) + (1-y_i) \log(1-\mathcal{M}_\theta(x_i))]
\end{equation}

The output of this phase is a performance-optimized model $\mathcal{M}_{\theta_{po}}(x)$ that achieves optimal predictive accuracy on the given healthcare task.

\subsection{Fairness Optimization}
In the second phase, we transfer the performance-optimized model $\mathcal{M}_{\theta_{po}}(x)$ to serve as the starting point for fairness optimization. Our goal is to improve fairness across multiple demographic groups while maintaining the model's predictive performance. The multi-objective optimization problem is formulated as:
\begin{equation}
\min_{\boldsymbol{\theta}} \ \mathcal{L}(\boldsymbol{\theta}) = \left\{ \ell_0(\boldsymbol{\theta}), \ell_1(\boldsymbol{\theta}), \ell_2(\boldsymbol{\theta}), \dots, \ell_n(\boldsymbol{\theta}) \right\}
\end{equation}
where $\ell_i ~ \forall i = 0,...n$ are the $n$ different objectives~\cite{valdivia2021fair}. $\mathcal{L}(\theta)$ is conceptualized as a composite objective function comprising multiple loss components. Each loss component, denoted as $\ell_i(\theta)$, represents a distinct optimization target for the machine learning model. To address this multi-objective optimization problem, we propose and investigate two distinct strategies: sequential and simultaneous. These strategies offer different approaches to balancing the various objectives $\ell_i(\theta)$, each with its own advantages and trade-offs in the context of fairness optimization for multiple demographic attributes in healthcare AI.
\subsubsection{Sequential Strategy}

In the sequential approach, as the Algorithm~\ref{alg:sequential_fairness_multi} presents, we optimize for each fairness attribute one at a time, starting from the performance-optimized model. The process can be described as:

\begin{align}
\mathcal{L}^{Seq}_1 &= \min_{\theta} (\Omega_0(\theta), \ell_1(\theta)) \\
\mathcal{L}^{Seq}_2 &= \min_{\theta} (\Omega_0(\theta), \Omega_1(\theta), \ell_2(\theta)) \\
&\vdots \\
\mathcal{L}^{Seq}_n &= \min_{\theta} (\Omega_0(\theta), \sum_{k=1}^{n-1} \Omega_k(\theta), \ell_n(\theta))
\end{align}

Here, to maintain both predictive performance and fairness improvements, we introduce a versatile regularization penalty. The performance penalty $\Omega_0(\theta)$ and the fairness penalties $\Omega_k(\theta)$ for $k = 1, 2, \dots, n-1$ are regularization penalties that prevent degradation of performance and fairness achieved in previous steps. They are defined as:

\begin{equation}
\Omega_{i}(\theta, \phi) = \mathbb{I}\{\phi(\mathcal{M}^{t-1}_{\theta}) - \epsilon \leq \phi(\mathcal{M}^{t}_{\theta}) \leq \phi(\mathcal{M}^{t-1}_{\theta}) + \epsilon\}
\end{equation}

where $\phi$ represents the metric of interest (either performance or fairness), $\mathcal{M}^{t-1}_{\theta}$ is the reference model (performance-optimized for $\Omega_0$ or intermediate fairness-optimized for $\Omega_i$), $\mathcal{M}^{t}_{\theta}$ is the current model, and $\epsilon$ is a tolerance parameter. This formulation allows for both performance regularization (when $\phi$ is a performance metric) and fairness regularization (when $\phi$ is a fairness metric for previously optimized attributes), ensuring that subsequent optimization steps do not significantly degrade earlier achievements in performance or fairness. $\ell_i(\theta)$ is the fairness loss for the current attribute being optimized, which is the weighted sum of the TPR difference and FPR difference as follows:
\begin{equation}
\ell_{i} = \frac{1}{2} \cdot (TPR_{a} - TPR_{b})^2 + \frac{1}{2} \cdot (FPR_{a} - FPR_{b})^2
\end{equation}

where $TPR_z = P(\hat{Y} = 1 | Y = 1, Z = z)$ and $FPR_z = P(\hat{Y} = 1 | Y = 0, Z = z)$ for groups $z \in \{a, b\}$ of sensitive attribute $Z$.

The indicator functions used to compute $TPR$ and $FPR$ are non-differentiable, which complicates gradient-based optimization. To overcome this, we approximate the 0-1 indicator function with a sigmoid function, defined as:
\begin{equation}
\sigma(x) = \frac{1}{1 + e^{-kx}}
\end{equation}
where $x$ represents the model’s output logits, and  $k$  is a hyperparameter that controls the steepness of the sigmoid curve. 

This sequential process results in a multi-attribute fairness-optimized model $\mathcal{M}_{\theta_{fo}}^{seq}$, where each step builds upon the fairness improvements of the previous steps while attempting to maintain performance.

\subsubsection{Simultaneous Strategy}

In the simultaneous approach, as the Algorithm~\ref{alg:simultaneous_fairness_multi} shows, we optimize for all fairness attributes simultaneously, balancing the trade-offs between different fairness objectives and performance in a single optimization step:

\begin{equation}
\mathcal{L}^{Sim} = \min_{\theta} (\Omega_0(\theta), \sum_{i=1}^{n} \ell_i(\theta))
\end{equation}

This approach directly optimizes the composite loss function, considering all fairness attributes simultaneously. The resulting model is denoted as $\mathcal{M}_{\theta_{fo}}^{sim}$.


\begin{algorithm}
\caption{Sequential Strategy for Multi-attribute Fairness Optimization}
\label{alg:sequential_fairness_multi}
\SetKwInOut{Input}{Input}\SetKwInOut{Output}{Output}
\SetKwComment{comment}{\#}{}
\Input{Training Samples $\mathcal{D}_{Z, Y}$, \\
       Sensitive attributes sets $\{Z_1, Z_2, ..., Z_n\}$, \\
       EOD thresholds $\{\zeta_1, \zeta_2, ..., \zeta_n\}$, \\
       Number of steps $T$, \\
       Performance-optimized model $\mathcal{M}_{\theta_{po}}$, \\
       Penalty Term $\Omega$
}
\Output{Multi-attribute Fairness-optimized Model: $\mathcal{M}^{Seq}_{\theta_{fo}}$ }
\BlankLine
Initialize $\mathcal{M}^{Seq}_{\theta_{fo}} \gets \mathcal{M}_{\theta_{po}}$, $\Omega_{total} \gets \Omega_{perf}$\;
\For{$i=1 \textbf{ to } n$}{
    $minEOD_i \gets \infty$\;
    \For{$t=1 \textbf{ to } T$}{
        $\mathcal{L}^{Seq}_i \gets \{\ell_{i}(\mathcal{D}_{Z, Y}, \theta), \Omega_{total}(\theta, \phi)\}$\;
        $\theta^t \gets \text{argmin}_\theta \, \mathcal{L}(\theta, \mathcal{D}_{Z, Y})$\;
        $eod_i \gets \phi(\mathcal{M}_{\theta^t}(\mathcal{D}_{Z, Y}), Z_i)$\;
        \If{$eod_i \le \zeta_i \land eod_i \le minEOD_i$}{
            $minEOD_i \gets eod_i$\;
            $\mathcal{M}^{Seq}_{\theta_{fo}} \gets \mathcal{M}_{\theta^t}$\;
        }
        \If{$eod_i \ge \zeta_i \land minEOD_i \neq \infty$}{
            \textbf{break}\;
        }
    }
    $\Omega_{total} \gets \Omega_{total} + \Omega_{i}$\;
}
\end{algorithm}

\begin{algorithm}
\caption{Simultaneous Strategy for Multi-attribute Fairness Optimization}
\label{alg:simultaneous_fairness_multi}
\SetKwInOut{Input}{Input}\SetKwInOut{Output}{Output}
\SetKwComment{comment}{\#}{}
\Input{Training Samples $\mathcal{D}_{Z, Y}$, \\
       Sensitive attributes sets $\{Z_1, Z_2, ..., Z_n\}$, \\
       EOD thresholds $\{\zeta_1, \zeta_2, ..., \zeta_n\}$, \\
       Number of steps $T$, \\
       Performance-optimized model $\mathcal{M}_{\theta_{po}}$
}
\Output{Multi-attribute Fairness-optimized Model: $\mathcal{M}^{Sim}_{\theta_{fo}}$ }
\BlankLine
 
Initialize $\mathcal{M}^{Sim}_{\theta_{fo}} \gets \mathcal{M}_{\theta_{po}}$, $minEOD_{total} \gets \infty$\;
\For{$t=1 \textbf{ to } T$}{
    $\theta^t \gets \text{argmin}_\theta \, \mathcal{L}^{Sim}(\theta, \mathcal{D}_{Z, Y}, \Omega_{perf})$\;
    $eod_{total} \gets 0$\;
    $fair\_all \gets \textbf{true}$\;
    \For{$i=1 \textbf{ to } n$}{
        $eod_i \gets \phi(\mathcal{M}_{\theta^t}(\mathcal{D}_{Z, Y}), Z_i)$\;
        $eod_{total} \gets eod_{total} + eod_i$\;
        \If{$eod_i > \zeta_i$}{
            $fair\_all \gets \textbf{false}$\;
        }
    }
    \If{$fair\_all \land eod_{total} < minEOD_{total}$}{
        $minEOD_{total} \gets eod_{total}$\;
        $\mathcal{M}^{Sim}_{\theta_{fo}} \gets \mathcal{M}_{\theta^t}$\;
    }
    \If{$\lnot fair\_all \land \mathcal{M}^{Sim}_{\theta_{fo}} \neq \mathcal{M}_0$}{
        \textbf{break}\;
    }
}
\end{algorithm}

\section{Experiments}
\subsection{Dataset}
We evaluate our proposed method on two real-world healthcare datasets, stratifying each dataset into distinct demographic subsets, delineated by protected attributes such as sexual (male/female) and racial identity (Caucasian/Non-Caucasian American). The distribution of target variables across these sensitive attributes, encompassing both negative and positive classes, is detailed in Tables~\ref{tab:dataset-characteristic-sud} and \ref{tab:dataset-characteristic-sepsis}. We utilized an 80\%-20\% split for training and testing sets, respectively. This split was stratified to maintain the distribution of sensitive attributes and outcome variables across all sets. The random seed was set to ensure reproducibility.

\textbf{The Substance Use Disorder (SUD) dataset }: This dataset originates from the Hazelden Betty Ford Foundation (HBFF) electronic health records(EHR)~\cite{liang_developing_2021}. It includes demographic information, socioeconomic variables, encounter-specific data, diagnosis-related variables, and responses to clinical questionnaires. Uniquely, it contains not only objective clinical measurements but also patient responses to questionnaires administered during treatment, including the American Society of Addiction Medicine (ASAM) Criteria, which measure substance use severity across six dimensions~\cite{liang_developing_2021}. The dataset comprises 10,673 instances after preprocessing. The task is to predict failure to complete treatment.

\textbf{The Sepsis dataset }: This dataset is derived from the MIMIC-IV database~\cite{johnson_mimic-iv_nodate}, which contains critical care records from Beth Israel Deaconess Medical Center's ICUs, focusing on patients diagnosed with sepsis. The final dataset includes demographic information, vital signs, and clinical scores. The dataset includes 9,349 instances after preprocessing. The target variable is patient mortality. 

\begin{table}[htbp]
\centering
\caption{SUD Dataset distribution of patients by sensitive attributes and class label}
\label{tab:dataset-characteristic-sud}
\resizebox{\columnwidth}{!}{%
\begin{tabular}{@{}lllll@{}}
\toprule
Characteristic                        & Negative Class (9,149) & Positive Class (1,524) \\ \midrule
Race                     &              &                          \\
\text{    }Caucasian               & 8,230 (90\%) & 1,341 (88\%)                   \\
\text{    }Non-Caucasian          & 919 (10\%)   & 183 (12\%)                     \\ 
Sex                      &              &               \\
 \text{    }Male                     & 5,824 (64\%) & 1,062 (70\%)                   \\
\text{    }Female                   & 3,325 (36\%) & 462 (30\%)                     \\\bottomrule
\end{tabular}%
}
\end{table}

\begin{table}[htbp]
\centering
\caption{Sepsis Dataset distribution of patients by sensitive attributes and class label}
\label{tab:dataset-characteristic-sepsis}
\resizebox{\columnwidth}{!}{%
\begin{tabular}{@{}lllll@{}}
\toprule
Characteristic                        & Negative Class (7,806) & Positive Class (1,543) \\ \midrule
Race                     &              &                          \\
\text{    }Caucasian               & 6,546 (83.9\%) & 1,251 (81.1\%)                   \\
\text{    }Non-Caucasian          & 1,260 (16.1\%)   & 292 (18.9\%)                     \\ 
Sex                      &              &               \\
 \text{    }Male                     & 4,496 (57.6\%) & 875 (56.7\%)                   \\
\text{    }Female                   & 3,310 (42.4\%) & 668 (43.3\%)                     \\\bottomrule
\end{tabular}%
}
\end{table}

\subsection{Baseline Methods} 
We compare the proposed method with two baseline methods, including:
    \begin{enumerate}
        \item \textit{Adversarial Debiasing}\cite{zhang_mitigating_2018}: reduces statistical parity by introducing an adversary to predict the sensitive attribute using the predicted outcome obtained from a predictor.
        
        \item \textit{Reduction Method}\cite{agarwal_reductions_2018}: convert fair classification into a series of cost-sensitive classification problems, solving them by generating a randomized classifier that has the lowest empirical error under the specified constraints, such as Demographic Parity and Equalized Odds. For a fair comparison, we evaluate the reduction method with the equalized odds constraint.
        
    \end{enumerate}
\subsection{Implementation Details} 
The Adversarial Debiasing and Reduction Methods were implemented using the IBM AIF360 package\footnote{The code and tutorial for AI Fairness 360 package can be found at \href{https://github.com/Trusted-AI/AIF360}{AIF360}}, a comprehensive toolkit for fairness-aware machine learning. As for parameter settings, we use the default number of prototypes as described in the implementation provided by IBM AIF360 to ensure reproducibility and fair comparison. Our proposed approach with two strategies for multi-attribute fairness: (1)Sequential optimization: Fairness optimization applied sequentially to each sensitive attribute (Algorithm~\ref{alg:sequential_fairness_multi}), (2)Simultaneous optimization: Fairness optimization applied simultaneously to all sensitive attributes (Algorithm~\ref{alg:simultaneous_fairness_multi})

\subsection{Model and Parameter Settings}
Due to high transparency and controllability, logistic regression was chosen as the base classifier for all methods to ensure fair comparison. The original paper on two baseline methods also applied logistic regression as the classifier. We tune the learning rate as 0.001 for baseline methods and our method. All learnable model parameters are optimized with Adam optimizer\cite{kingma2014adam}.  A batch size of 1,000 was used for training. All experiments were repeated 5 times with different initializations with random seeds to enhance the robustness of the results. The performance metrics and fairness measures were averaged across these runs, and standard deviations were computed to assess the stability of the results.

\subsection{Evaluation Metrics}
We evaluate our models with the following metrics:
\subsubsection{Area Under the Receiver Operating Characteristic Curve (AUROC)}
AUROC measures the model's ability to distinguish between classes. It is calculated as the area under the Receiver Operating Characteristic (ROC) curve, which plots the True Positive Rate (TPR) against the False Positive Rate (FPR) at various threshold settings.

\subsubsection{Sensitivity and Specificity}
Sensitivity measures the proportion of actual positive cases correctly identified:

\begin{equation}
    \text{Sensitivity} = \frac{\text{True Positives}}{\text{True Positives} + \text{False Negatives}}
\end{equation}

Specificity measures the proportion of actual negative cases correctly identified:

\begin{equation}
    \text{Specificity} = \frac{\text{True Negatives}}{\text{True Negatives} + \text{False Positives}}
\end{equation}

\subsubsection{Equalized Odds Disparity (EOD)}
Equalized Odds Disparity (EOD) quantifies the fairness of the model with respect to sensitive attributes~\cite{hardt2016equality}. It is calculated as the average of the absolute differences in TPR and FPR between groups defined by a sensitive attribute:

\begin{equation}
    \text{EOD} = \frac{1}{2}(|\text{TPR}_a - \text{TPR}_b| + |\text{FPR}_a - \text{FPR}_b|)
\end{equation}

where $a$ and $b$ represent two groups defined by a sensitive attribute (e.g., male and female for sex, or Caucasian and non-Caucasian for race). 

A lower EOD indicates better fairness, with 0 representing perfect equality of odds. For our multi-attribute fairness scenarios, we calculate separate EOD values for each sensitive attribute (Race EOD and Sex EOD) to assess fairness across different demographic dimensions.

\begin{table*}[htbp]
\centering
\caption{Model performance and fairness - SUD}
\label{tab:performances_and_fairness_SUD}
\resizebox{\textwidth}{!}{%
\begin{tabular}{lccccccc}
\toprule
\multirow{2}{*}{\makecell[c]{Fair\\Method}} & \multirow{2}{*}{Model} & \multirow{2}{*}{AUROC} & \multirow{2}{*}{Sensitivity} & \multirow{2}{*}{Specificity} & \multicolumn{2}{c}{EOD} \\
\cmidrule{6-7}
& & & & & Race & Sex \\
\midrule
None & Best Performing Model & 0.8640 & 0.8092 & 0.7977 & 0.0513 & 0.0574 \\ 
\midrule
\multirow{3}{*}{Adversarial} & Race-Fair Model & 0.8635 & 0.7996 & 0.8015 & 0.0402 & 0.0652 \\
& Sex-Fair Model & 0.8602 & 0.7925 & 0.8125 & 0.0612 & 0.0392 \\
\cmidrule{2-7}
& Multi-Fair Model & 0.8615 & 0.7962 & 0.8082 & 0.0395 & 0.0455 \\
\cmidrule{1-7}
\multirow{3}{*}{Reduction} & Race-Fair Model & 0.8472 & 0.7812 & 0.7889 & 0.0205 & 0.0592 \\
& Sex-Fair Model & 0.8489 & 0.7725 & 0.7969 & 0.0575 & 0.0212 \\
\cmidrule{2-7}
& Multi-Fair Model & 0.8265 & 0.7862 & 0.7724 & 0.0262 & 0.0315 \\
\cmidrule{1-7}
\multirow{3}{*}{Our Method} & Race-Fair Model & 0.8633 & 0.7982 & 0.8039 & 0.0226 & 0.0607 \\ 
& Sex-Fair Model & 0.8585 & 0.7895 & 0.8119 & 0.0596 & 0.0258 \\
\cmidrule{2-7}
& Multi-Fair Model & 0.8613 & 0.7654 & 0.8199 & 0.0274 & 0.0346 \\
\bottomrule
\end{tabular}
}
\end{table*}

\begin{table*}[htbp]
\centering
\caption{Model performance and fairness - Sepsis}
\label{tab:performances_and_fairness_Sepsis}
\resizebox{\textwidth}{!}{%
\begin{tabular}{lccccccc}
\toprule
\multirow{2}{*}{\makecell[c]{Fair\\Method}} & \multirow{2}{*}{Model} & \multirow{2}{*}{AUROC} & \multirow{2}{*}{Sensitivity} & \multirow{2}{*}{Specificity} & \multicolumn{2}{c}{EOD} \\
\cmidrule{6-7}
& & & & & Race & Sex \\
\midrule
None & Best Performing Model & 0.7467 & 0.7149 & 0.6712 & 0.0753 & 0.0351 \\ 
\midrule
\multirow{3}{*}{Adversarial} & Race-Fair Model & 0.7312 & 0.6932 & 0.6592 & 0.0468 & 0.0455 \\
& Sex-Fair Model & 0.7465 & 0.7092 & 0.6732 & 0.0862 & 0.0248 \\
\cmidrule{2-7}
& Multi-Fair Model & 0.7385 & 0.7015 & 0.6645 & 0.0492 & 0.0395 \\

\cmidrule{1-7}
\multirow{3}{*}{Reduction} & Race-Fair Model & 0.7185 & 0.6892 & 0.6562 & 0.0212 & 0.0375 \\
& Sex-Fair Model & 0.7232 & 0.6865 & 0.6685 & 0.0838 & 0.0141 \\
\cmidrule{2-7}
& Multi-Fair Model & 0.7052 & 0.6775 & 0.6602 & 0.0245 & 0.0282 \\

\cmidrule{1-7}
\multirow{3}{*}{Our Method} & Race-Fair Model & 0.7306 & 0.6911 & 0.6584 & 0.0215 & 0.0388 \\ 
& Sex-Fair Model & 0.7453 & 0.7084 & 0.6704 & 0.0841 & 0.0143 \\
\cmidrule{2-7}
& Multi-Fair Model & 0.7375 & 0.6995 & 0.6632 & 0.0265 & 0.0312 \\
\bottomrule
\end{tabular}
}
\end{table*}

\section{Results and Discussions}
Tables \ref{tab:performances_and_fairness_SUD} and \ref{tab:performances_and_fairness_Sepsis} present the impact of different fairness optimization methods on our models' predictive performance and fairness metrics for the Substance Use Disorder (SUD) and Sepsis datasets. We compare three approaches: Adversarial Debiasing\cite{zhang_mitigating_2018}, Reduction-based method\cite{agarwal_reductions_2018}, and our proposed method. For each method, we present results for models optimized for race fairness, sex fairness, and multi-attribute fairness. Note that all multi-attribute fairness optimization in these two tables adopts a simultaneous strategy, that is, optimizing multiple different attributes at the same time.

\subsection{Single-attribute Fairness}
Our experiments reveal distinct trade-offs between performance and fairness across different methods. The adversarial method achieves competitive predictive performance but shows limitations in fairness improvement. For instance, in the SUD dataset, while maintaining high AUROC (0.8635 for Race-Fair Model), it shows larger fairness disparities (Race EOD of 0.0402 compared to our method's 0.0226). Similarly, for sex fairness, while achieving an AUROC of 0.8602, it results in a Sex EOD of 0.0392, higher than our method's 0.0258.

The Reduction-based method, conversely, achieves better fairness metrics but at a significant cost to predictive performance. In the SUD dataset, while achieving a Race EOD of 0.0205, its Race-Fair Model shows notably lower AUROC (0.8472) compared to our method (0.8633). This pattern is consistently observed in the Sepsis dataset, where the Reduction method's Race-Fair Model achieves a Race EOD of 0.0212 but with an AUROC of only 0.7185, compared to our method's AUROC of 0.7306.

Our proposed method demonstrates a more balanced trade-off between performance and fairness. It maintains competitive AUROC scores (0.8633 for Race-Fair Model in SUD dataset) while achieving significant fairness improvements (Race EOD of 0.0226). This balanced performance is consistent across both datasets and both protected attributes, suggesting that our method effectively addresses the challenging task of maintaining predictive performance while improving fairness.

These results highlight the importance of considering both performance and fairness metrics when evaluating fairness optimization methods. While some methods may excel in one aspect, achieving a balanced improvement in both dimensions is crucial for practical applications in healthcare settings.

\subsection{Multi-attribute Fairness}
When examining multi-attribute fairness optimization, we observe distinct patterns across the three methods. For both SUD and Sepsis datasets, each method exhibits different characteristics in balancing performance and fairness across multiple attributes simultaneously.

The Adversarial method's multi-fair model maintains high predictive performance (AUROC of 0.8615 for SUD and 0.7385 for Sepsis) but shows limitations in achieving balanced fairness improvements. In the SUD dataset, its Race EOD (0.0395) and Sex EOD (0.0455) remain higher than both single-attribute optimization results, suggesting difficulties in simultaneously addressing multiple fairness objectives.

The Reduction method shows the opposite trend. Its multi-fair model achieves better fairness metrics (Race EOD of 0.0262 and Sex EOD of 0.0315 for SUD) but suffers from substantial performance degradation (AUROC of 0.8265 for SUD and 0.7052 for Sepsis). This significant drop in predictive performance could limit its practical applicability in healthcare settings where maintaining high accuracy is crucial.

Our method demonstrates a more balanced approach to multi-attribute fairness. For the SUD dataset, our multi-fair model achieves an AUROC of 0.8613 while maintaining competitive fairness metrics (Race EOD of 0.0274 and Sex EOD of 0.0346). Similarly, in the Sepsis dataset, our method achieves an AUROC of 0.7375 with Race EOD of 0.0265 and Sex EOD of 0.0312. These results suggest that our method can effectively optimize for multiple fairness constraints while preserving predictive performance.

Notably, all methods show some degradation in performance when optimizing for multiple attributes compared to single-attribute optimization. However, our method exhibits the most stable performance across both single and multi-attribute scenarios. This stability is particularly important in healthcare applications where maintaining consistent model performance across different fairness objectives is essential.

\subsection{Different Strategies for Proposed Method}
Tables \ref{tab:different_implementation_SUD} and \ref{tab:different_implementation_Sepsis} present the results of different strategies of our multi-attribute fairness optimization method on the SUD and Sepsis datasets, respectively. We compare the sequential strategy (e.g., Sequential(Race, Sex)), where fairness is optimized for one attribute followed by the other, with the simultaneous strategy (Simultaneous Race \& Sex) that optimizes for both attributes at the same time. Our analysis reveals several key insights into the effectiveness and characteristics of these different strategies.

\subsubsection{Attribute Prioritization in Sequential Strategy}

In the sequential approach, the attribute optimized first tends to have better fairness outcomes. For example, in the SUD dataset, when race fairness is optimized first, we see a better optimized Race EOD (0.0250) compared to Sex EOD (0.0380). Conversely, when sex fairness is prioritized, the Sex EOD (0.0290) is better optimized than the Race EOD (0.0370). The simultaneous approach, in contrast, achieves a more balanced improvement with Race EOD at 0.0274 and Sex EOD at 0.0346.

A similar pattern emerges in the Sepsis dataset. The Sequential(Race, Sex) sequence results in a better optimized Race EOD (0.0281) compared to Sex EOD (0.0258), while the Sequential(Sex, Race) sequence yields a better optimized Sex EOD (0.0208) compared to Race EOD (0.0320). Once again, the simultaneous approach shows more balanced improvements with Race EOD at 0.0307 and Sex EOD at 0.0195.

This consistent pattern suggests that the initial optimization step in the sequential approach tends to favor the first attribute, which persists even after the second optimization step. The simultaneous approach avoids this favor and achieves a more equitable distribution of fairness improvements. This phenomenon is consistently observed in both datasets.

\subsubsection{Performance-Fairness Trade-offs}

The different strategies show varying trade-offs between predictive performance and fairness. In the SUD dataset, the sequential strategy maintains slightly higher AUROC (0.8607 and 0.8631) compared to the simultaneous approach (0.8613). Similarly, for the Sepsis dataset, the sequential strategy shows marginally higher AUROC (0.7353 and 0.7346) than the simultaneous method (0.7335). However, these small performance gains come at the cost of less balanced fairness improvements across attributes.

\begin{table*}[htbp]
\centering
\caption{Different Fairness Consideration - SUD}
\label{tab:different_implementation_SUD}
\resizebox{\textwidth}{!}{%
\begin{tabular}{llcccccc}
\toprule
Fairness Consideration & Fairness Strategy & AUROC & Sensitivity & Specificity & Race EOD & Sex EOD \\
\midrule
\multirow{3}{*}{Multi-fair}
 & Sequential(Race, Sex) & 0.8607 & 0.8004 & 0.7820 & \textbf{0.0250} & 0.0380\\ 
 & Sequential(Sex, Race) & 0.8631 & 0.7917 & 0.7955 & 0.0370 & \textbf{0.0290} \\ 
\cmidrule(l){2-7}
 & Simultaneous(Race \& Sex) & 0.8613 & 0.7654 & 0.8199 & \textbf{0.0274} & \textbf{0.0346} \\ 
\bottomrule
\end{tabular}
}
\end{table*}

\begin{table*}[htbp]
\centering
\caption{Different Fairness Consideration - Sepsis}
\label{tab:different_implementation_Sepsis}
\resizebox{\textwidth}{!}{%
\begin{tabular}{llcccccc}
\toprule
Fairness Consideration & Fairness Strategy & AUROC & Sensitivity & Specificity & Race EOD & Sex EOD \\
\midrule
\multirow{3}{*}{Multi-fair}
 & Sequential(Race, Sex) & 0.7353 & 0.7183 & 0.6609 & \textbf{0.0281} & 0.0258\\ 
 & Sequential(Sex, Race) & 0.7346 & 0.6981 & 0.6784 & 0.0320 & \textbf{0.0208} \\ 
\cmidrule(l){2-7}
 & Simultaneous(Race \& Sex) & 0.7335 & 0.7322 & 0.6452 & \textbf{0.0307} & \textbf{0.0195} \\ 
\bottomrule
\end{tabular}
}
\end{table*}

\subsection{Discussion}
\textbf{Single-attribute fairness optimization methods, while effectively optimizing fairness for the target attribute, may inadvertently increase disparities in other sensitive attributes.} In the context of Healthcare AI, this situation raises significant ethical concerns. Healthcare systems serve diverse populations with intersecting demographic characteristics, and biased AI models could exacerbate existing health disparities or create new ones. Our experimental results clearly demonstrate this phenomenon across different fairness optimization methods. For instance, in the SUD dataset, the Adversarial method's Race-Fair model reduces Race EOD from 0.0513 to 0.0402, but simultaneously increases Sex EOD from 0.0574 to 0.0652. Similarly, its Sex-Fair model improves Sex EOD but leads to increased Race EOD. This observation aligns with previous findings by Chen et al.~\cite{chen2024fairness}, who demonstrated that some fairness improvement methods can lead to decreased fairness regarding unconsidered protected attributes to a large extent. The Reduction method shows similar trade-offs, albeit with different characteristics - while achieving better fairness for the targeted attribute, it shows significant performance degradation that could impact clinical reliability. For healthcare scenarios, a model that achieves fairness for sensitive attribute A but neglects sensitive attribute B differences might lead to misdiagnoses or inappropriate treatment recommendations for certain subgroups, potentially compromising patient safety and outcomes.

\textbf{Sequential strategy of fairness optimization tends to prioritize the first-optimized attribute, resulting in uneven fairness improvements. In contrast, simultaneous optimization achieves more balanced fairness enhancements across attributes.} While sequential approaches may offer slight advantages in overall predictive performance (AUROC), the simultaneous method provides a more equitable solution for multi-attribute fairness. In the context of healthcare, the choice between these approaches could have significant implications for clinical decision-making and patient outcomes. For diseases with known disparities in certain demographic groups, prioritizing fairness for those attributes through sequential optimization could be beneficial. However, for conditions where the interplay of multiple demographic factors is less understood, the balanced approach of simultaneous optimization might be more appropriate. Ultimately, the decision between sequential and simultaneous fairness optimization in healthcare AI should be guided by the specific clinical context, the potential impact on patient outcomes, and the ethical considerations of fairness in the given healthcare scenario. 

These findings underscore the importance of carefully considering strategies when addressing multiple fairness concerns in AI systems, particularly in sensitive domains such as healthcare.

\section{Conclusion and Future Work}
In this study, we presented an approach to addressing multi-demographic fairness in healthcare AI systems through transfer learning. Our method demonstrates the ability to significantly reduce Equalized Odds Disparity (EOD) for multiple demographic attributes while largely maintaining predictive performance across two critical healthcare domains: Substance Use Disorder (SUD) treatment completion prediction and sepsis mortality prediction. Specifically, our experiments showed that sequential strategy tends to favor the first-optimized attribute, while simultaneous strategy achieves more balanced fairness improvements. 

Importantly, we observed that single-fairness optimization methods effectively optimize fairness for the target attribute but may inadvertently increase disparities in other sensitive attributes. In contrast, our multi-attribute fairness optimization approach addresses this issue by providing a more equitable improvement across all considered attributes. These findings are crucial for ensuring equitable care and developing strategies that address multiple fairness concerns in healthcare AI.

While our current work provides valuable insights, several avenues for future research remain open. Future efforts should explore more sophisticated techniques for balancing multiple fairness objectives. This could involve advanced multi-objective optimization algorithms or novel loss function designs that better capture the complexities of fairness in healthcare contexts. How to extend our fairness optimization method to multi-class population groups will also be studied in future work to ensure that it can address unfairness issues in more complex real-world healthcare data. Additionally, as healthcare data becomes increasingly diverse, incorporating multi-modal inputs presents both challenges and opportunities for fairness-aware AI. Future research should investigate how our fairness optimization approach can be extended to multi-modal models, ensuring fairness across varied data types and sources such as electronic health records, medical imaging, and genomic data. 

By addressing multi-attribute fairness and maintaining high predictive performance, our work moves us closer to developing AI systems that can be reliably and ethically deployed in real-world healthcare settings. Promoting fairness across multiple demographic attributes not only enhances the ethical standing of AI applications but also contributes to reducing health disparities and improving patient outcomes.

\section*{Acknowledgment}
This work was supported in part by the National Science Foundation under the Grants IIS-1741306 and IIS-2235548, and by the Department of Defense under the Grant DoD W91XWH-05-1-023.  This material is based upon work supported by (while serving at) the National Science Foundation.  Any opinions, findings, conclusions, or recommendations expressed in this material are those of the author(s) and do not necessarily reflect the views of the National Science Foundation.

\bibliographystyle{IEEEtran}

\bibliography{main}

\end{document}